\documentclass[11pt,a4paper,dvipsnames]{article}
\usepackage[hyperref]{emnlp2020}
\usepackage[utf8]{inputenc}
\usepackage{times}
\usepackage{latexsym}
\usepackage{booktabs}
\usepackage{tikz}
\usepackage{adjustbox}
\usepackage{multirow}
\usepackage{collcell}
\usepackage{xstring}
\usepackage[all]{nowidow}

\usepackage{microtype}

\usepackage{float}
\usepackage{newfloat}
\usepackage{caption}
\DeclareFloatingEnvironment[fileext=frm,name=Frame]{myfloat}

\usepackage{url}
\usepackage{todonotes}

\usepackage{multicol}
\newcounter{notecounter}
\newcommand{\enotesoff}{\long\gdef\enote##1##2{}}
\newcommand{\enoteson}{\long\gdef\enote##1##2{{
			\stepcounter{notecounter}
			\large\bf
			\hspace{100cm}\arabic{notecounter} $<<<$ ##1: ##2
			$>>>$\hspace{1cm}}}}
\enoteson
\enotesoff

\aclfinalcopy 
\def\aclpaperid{1416} 

\def\aclpaperid{1416} 

\newcommand{\myparagraph}[1]{\paragraph{#1}}

\def\tablescale{0.94}
\newcommand{\cls}{\texttt{\small [cls]}}

\title{On the Language Neutrality of Pre-trained Multilingual Representations}

 \author{Jindřich Libovický$^1$ \and Rudolf Rosa$^2$ \and Alexander Fraser$^1$ \\
 \\
 $^1$Center for Information and Language Processing, LMU Munich, Germany \\
 $^2$Faculty of Mathematics and Physics, Charles University, Prague, Czech Republic \\
 \texttt{\{libovicky, fraser\}@cis.lmu.de rosa@ufal.mff.cuni.cz}
     }

\date{}

\begin{document}

\maketitle

\begin{abstract}
Multilingual contextual embeddings, such as multilingual BERT and XLM-RoBERTa,
    have proved useful for many multi-lingual tasks.
Previous work probed the cross-linguality of the representations indirectly
    using zero-shot transfer learning on morphological and syntactic tasks.
We instead investigate the language-neutrality of multilingual contextual
    embeddings directly and with respect to lexical semantics.
Our results show that contextual embeddings are more language-neutral and, in
    general, more informative than aligned static word-type embeddings, which
    are explicitly trained for language neutrality.
Contextual embeddings are still only moderately language-neutral by default, so
    we propose two simple methods for achieving stronger language neutrality:
    first, by unsupervised centering of the representation for each language
    and second, by fitting an explicit projection on small parallel data.
Besides, we show how to reach state-of-the-art accuracy on language
    identification and match the performance of statistical methods for word
    alignment of parallel sentences without using parallel data.

%
\end{abstract}

\section{Introduction}

Multilingual BERT (mBERT\@; \citealt{devlin2019bert}) gained popularity as a
contextual representation for many multilingual tasks, e.g., dependency parsing
\citep{kondratyuk2019udify,wang2019crosslingual}, cross-lingual natural
language inference (XNLI) or named-entity recognition (NER)
\citep{pires2019multilingual,wu2019beto,kudugunta2019investigating}. Recently,
a new pre-trained model, XLM-RoBERTa (XLM-R\@;
\citealt{conneau2019unsupervised}), claimed to outperform mBERT both on XNLI and
NER tasks. We also study DistilBERT \citep{sanh2019distilbert} applied to
mBERT, which promises to deliver comparable results to mBERT at a significantly
lower computational cost.

\citet{pires2019multilingual} present an exploratory paper showing that mBERT
can be used cross-lingually for zero-shot transfer in morphological and
syntactic tasks, at least for typologically similar languages. They also study
an interesting semantic task, sentence-retrieval, with promising initial
results. Their work leaves many open questions regarding how well the
cross-lingual mBERT representation captures lexical semantics, motivating our
work.

In this paper, we directly assess the cross-lingual properties of multilingual
representations on tasks where lexical semantics plays an important role and
present one unsuccessful and two successful methods for achieving better
language neutrality.

Multilingual capabilities of representations are often evaluated by zero-shot
transfer from the training language to a test language \citep{hu2020extreme,lian2020xglue}.
However, in such a setup, we can
never be sure if the probing model did not overfit for the original language,
as training is usually stopped when accuracy decreases
on a validation set from the same language (otherwise, it would not be zero-shot),
even when it would have been better to stop the training earlier. This
overfitting on the original language can pose a disadvantage for
information-richer representations.

To avoid such methodological issues, we select tasks that only involve a direct
comparison of the representations with no training: cross-lingual sentence
retrieval, word alignment (WA), and machine translation quality estimation (MT
QE).  Additionally, we explore how the language is represented in the
embeddings by training language ID classifiers and assessing how the
representation similarity corresponds to phylogenetic language families.

We find that contextual representations are more language-neutral than static
word embeddings which have been explicitly trained to represent matching words
similarly and can be used in a simple algorithm to reach state-of-the-art
results on word alignment. However, they also still strongly carry information
about the language identity, as demonstrated by a simple classifier trained on
mean-pooled contextual representations reaching state-of-the-art results on
language identification.

We show that the representations can be modified to be more language-neutral
with simple, straightforward setups: centering the representation for each
language or fitting explicit projections on small parallel data.

We further show that XLM-RoBERTa (XLM-R\@; \citealt{conneau2019unsupervised})
outperforms mBERT in sentence retrieval and MT QE while offering a similar
performance for language ID and WA\@.

\section{Related Work}

Multilingual representations, mostly mBERT, were already tested in a wide range
of tasks.  Often, the success of zero-shot transfer is implicitly considered to
be the primary measure of language neutrality of a representation.  Despite
many positive results, some findings in the literature are somewhat mixed,
indicating limited language neutrality.

Zero-shot learning abilities were examined by \citet{pires2019multilingual} on
NER and part-of-speech (POS) tagging, showing that the success strongly depends
on how typologically similar the languages are. Similarly, \citet{wu2019beto}
trained good multilingual models but struggled to achieve good results in the
zero-shot setup for POS tagging, NER, and XLNI\@.
\citet{ronnqvist2019multilingual} draw similar conclusions for
language-generation tasks.

\citet{wang2019crosslingual} succeeded in zero-shot dependency parsing but
required supervised projection trained on word-aligned parallel data. The
results of \citet{chi2020finding} on dependency parsing suggest that methods
like structural probing \citep{hewitt-manning-2019-structural} might be more
suitable for zero-shot transfer.

\citet{pires2019multilingual} also assessed mBERT on cross-lingual sentence
retrieval between three language pairs. They observed that if they subtract the
average difference between the embeddings from the target language
representation, the retrieval accuracy significantly increases. We
systematically study this idea in the later sections.

XTREME \citep{hu2020extreme} and XGLUE \citep{lian2020xglue}, two recently
introduced benchmarks for multilingual representation evaluation, assess
representations on a broader range of zero-shot transfer tasks that include
natural language inference \citep{conneau2018xnli} and question answering
\citep{artexe2019cross,lewis2019mlqa}. Their results show a clearly superior
performance of XLM-R compared to mBERT\@.

Many works clearly show that downstream task models can extract relevant
features from the multilingual representations
\citep{wu2019beto,kudugunta2019investigating,kondratyuk2019udify}. However,
they do not directly show language-neutrality, i.e., to what extent similar
phenomena are represented similarly across languages.  Thus, it is impossible
to say whether the representations are language-agnostic or contain some
implicit language identification. Our choice of evaluation tasks eliminates
this risk by directly comparing the representations.

\section{Centering Representations}

One way to achieve stronger language neutrality is by suppressing the language
identity, only keeping what encodes the sentence meaning. It can be achieved,
for instance, using an explicit projection. However, training such a projection
requires parallel data. Instead, we explore a simple unsupervised method:
representation centering.

Following \citet{pires2019multilingual}, we hypothesize that a sentence
representation in mBERT is additively composed of a language-specific
component, which identifies the language of the sentence, and a
language-neutral component, which captures the meaning of the sentence in a
language-independent way. We assume that the language-specific component is
similar across all sentences in the language.

We estimate the \textit{language centroid} as the mean of the representations
for a set of sentences in that language and subtract the language centroid from
the contextual embeddings. By doing this, we are trying to remove the
language-specific information from the representations by centering the
sentence representations in each language so that their average lies at the
origin of the vector space.

The intuition behind this is that within one language, certain phenomena (e.g.,
function words) would be very frequent, thus being quite prominent in the mean
of the representations for that language (but not for a different language),
while the phenomena that vary among sentences of the language (and thus
presumably carry most of the meaning) would get averaged out in the centroid.
We thus hypothesize that by subtracting the centroid, we remove the
language-specific features (without much loss of the meaning content), making
the meaning-bearing features more prominent.

We analyze the semantic properties of the original and the centered
representations on a range of probing tasks. For all tasks, we test all layers
of the model. We test both the \cls{} token vector and mean-pooled states for
tasks utilizing a single-vector sentence representation.

\section{Probing Tasks}

We employ five probing tasks to evaluate the language neutrality of the
representations.

The first two tasks analyze the contextual embeddings. The other three tasks
are cross-lingual NLP problems, all of which can be treated as a general task
of a cross-lingual estimation of word or sentence similarities. Supposing we
have sufficiently language-neutral representations, we can estimate these
similarities using the cosine distance of the representations; the performance
in these tasks can thus be viewed as a measure of the language-neutrality of
the representations.

Moreover, in addition to such an unsupervised approach, we can also utilize
actual training data for the tasks to further improve the performance of the
probes; this does not tell us much more about the representations themselves
but leads to a nice by-product of reaching state-of-the-art accuracies for two
of the tasks.

\myparagraph{Language Identification.} 
With a representation that captures all phenomena in a language-neutral way, it
should be difficult to determine what language the sentence is written in.
Unlike our other tasks, language ID requires fitting a classifier\@. We train a
linear classifier on top of a sentence representation.

\myparagraph{Language Similarity.} 
Previous work \citep{pires2019multilingual,wang2019crosslingual} shows that
models can be transferred better between more similar languages, suggesting
that similar languages tend to get similar representations. We quantify this
observation by V-measure between language families and hierarchical clustering
of the language centroids \citep{rosenberg2007vmeasure}. We cluster the
language centroids by their cosine distance using the Nearest Point Algorithm
and stop the clustering with a number of clusters equal to the number of
language families in the data.

\myparagraph{Parallel Sentence Retrieval.} 
For each sentence in a multi-parallel corpus, we compute the cosine distance of
its representation with representations of all sentences on the parallel side
of the corpus and select the sentence with the smallest distance.

Besides the plain and centered representations, we evaluate explicit projection
of the representations into the ``English space.'' We fit the projection by
minimizing the element-wise mean squared error between the representation of an
English sentence and a linear projection of the representation of its
translation.

\myparagraph{Word Alignment.} 
WA is the task of matching words which are translations of each other in
parallel sentences. WA is a key component of statistical machine translation
systems \citep{koehn2009smt}. While sentence retrieval could be done with
keyword spotting, computing bilingual WA requires resolving detailed
correspondence on the word level. Unsupervised statistical methods trained on
parallel corpora \citep{och2003systematic,dyer2013simple} still pose a strong
baseline for the task. In a work parallel to ours, \citet{sabet2020simalign}
present a more complex alternative way of leveraging contextual representations
for word alignment that outperforms the statistical methods.

For a pair of parallel sentences, we find the WA as a minimum weighted edge
cover of a bipartite graph. We create an edge for each potential alignment
link, weight it by the cosine distance of the token representations, and find
the WA as a minimum weighted edge cover of the resulting bipartite graph.
Unlike statistical methods, this does not require parallel data for training.

To make the algorithm prefer monotonic alignment, we add a distortion penalty
of $1/d$ to each edge where $d$ is the difference in the absolute positions of
the respective tokens in the sentence. We add the penalty with a weight that is
a hyper-parameter of the method estimated on a development set.

We keep the tokenization as provided in the word alignment dataset. In the
matching phase, we represent the tokens that get split into multiple subwords
as the average of the embeddings of the subwords.

Note that this algorithm is invariant to representation centering. Centering
the representation would shift all vectors by a constant. Therefore, all
weights would change by the same offset, not influencing the edge cover. We
evaluate WA using F$_1$ over sure and possible alignments in manually aligned
data.

\myparagraph{MT Quality Estimation.} 
MT QE assesses the quality of an MT system output without having access to a
reference translation. Semantic adequacy that we can estimate by comparing
representations of the source sentence and translation hypothesis can be a
strong indicator of the MT quality.
The standard evaluation metric is the Pearson correlation with the Translation
Error Rate (TER)---the number of edit operations a human translator would need
to do to correct the system output. QE is a more challenging task than the
previous ones because it requires capturing more subtle differences in meaning.

We evaluate how cosine distance of the representation of the source sentence
and of the MT output reflects the translation quality. In addition to plain and
centered representations, we also test trained bilingual projection and a fully
supervised regression trained on the shared task training data.

We use the same bilingual projection into English space fitted by linear
regression on the small parallel data used for sentence retrieval.

For the supervised regression, we use a multilayer perceptron directly
predicting the value of the translation error rate provided in the training
data.

Note that this task differs from reference-free MT evaluation
\citet[Task~3]{fonseca2019findings}, which is evaluated by computing the
correlation of the estimated value with human assessment of translation quality
based on reference sentences (available only to the annotators and not to the
evaluation metric). This task was also recently used for assessing the quality
of multilingual contextual representations
\citep{zhao-etal-2020-limitations,zhao2020inducing}.

\section{Probed Models}

\myparagraph{Aligned static word embeddings.} 

As a baseline in all our experiments, we use aligned static word
embeddings \citep{joulin2018loss}. Unlike hidden states of pre-trained Transformers, they do
not capture sentence context. However, they were explicitly trained to be
language-neutral with respect to lexical semantics. We represent sentences as
an average of the embeddings of the words.

\myparagraph{Multilingual BERT} \hspace{-2ex} \citep{devlin2019bert} is a deep
Transformer \citep{vaswani2017attention} encoder that is trained in a
multi-task learning setup, first, to be able to guess what words were
masked-out in the input and, second, to decide whether two sentences follow
each other in a coherent text.

We use a pre-trained mBERT model that was made public with the BERT
release.\footnote{https://github.com/google-research/bert} The model dimension
is 768, the hidden layer dimension 3072, self-attention uses 12 heads, the
model has 12 layers. It uses a vocabulary of 120k wordpieces shared for all
languages.

It is trained using a combination of a masked language model (MLM) objective
and sentence-adjacency objective. For the MLM objective, 15\% of input subwords
are masked out, and the model predicts the masked subwords. For the
sentence-adjacency objective, a special \cls{} token is prepended to the input.
The embedding corresponding to this token is used as an input to a classifier
predicting if the input sentences are adjacent.

Therefore, for models based on mBERT, we experiment both with \cls{} vector and
the \emph{mean-pooled} vector, i.e., average embeddings for the rest of the
tokens.

\myparagraph{UDify.} 
The UDify model \citep{kondratyuk2019udify} uses mBERT to train a single model
for dependency parsing and morphological analysis of 75 languages. During
training, mBERT is finetuned, which improves accuracy. Results on zero-shot
parsing suggest that the finetuning leads to better language neutrality with
respect to morphology and syntax.


\myparagraph{lng-free.} 
In this experiment, we try to make the representations more language-neutral by
removing the language identity from the model using an adversarial approach.
We continue training mBERT in a multi-task learning setup with the MLM
objective \citep{devlin2019bert} without the sentence adjacency objective,
i.e., the same way as XLM-R. It is trained jointly with adversarial language ID
classifiers \citep{elazar2018adversarial} using the same dataset as for the
language ID tasks. The classifier is separated from the rest of the model by a
gradient-reversal layer \citep{ganin2015unsupervised}, which negates the
gradients flowing from the classifier into the model. Intuitively, we can say
that the rest of the model is trying to fool the classifier, whereas the
classifier tries to improve.

\begin{table*}[t]

	\centering

    \scalebox{\tablescale}{%
    \begin{tabular}{lccccc}
        \toprule
        & mBERT & UDify & lng-free & Distil & XLM-R \\ \midrule

        \cls{}           & .935 & .938 & .796 & .953 & ---  \\
        \cls, cent.      & .867 & .851 & .337 & .826 & N/A  \\ \midrule
        mean-pool        & .960 & .959 & .951 & .953 & .950 \\
        mean-pool, cent. & .853 & .854 & .855 & .826 & .846 \\ \bottomrule

    \end{tabular}}
	\caption{Accuracy of language identification, values from the best-scoring
	layers.}\label{tab:lngid}

\end{table*}

\begin{figure*}[t]

    \centering
        \includegraphics{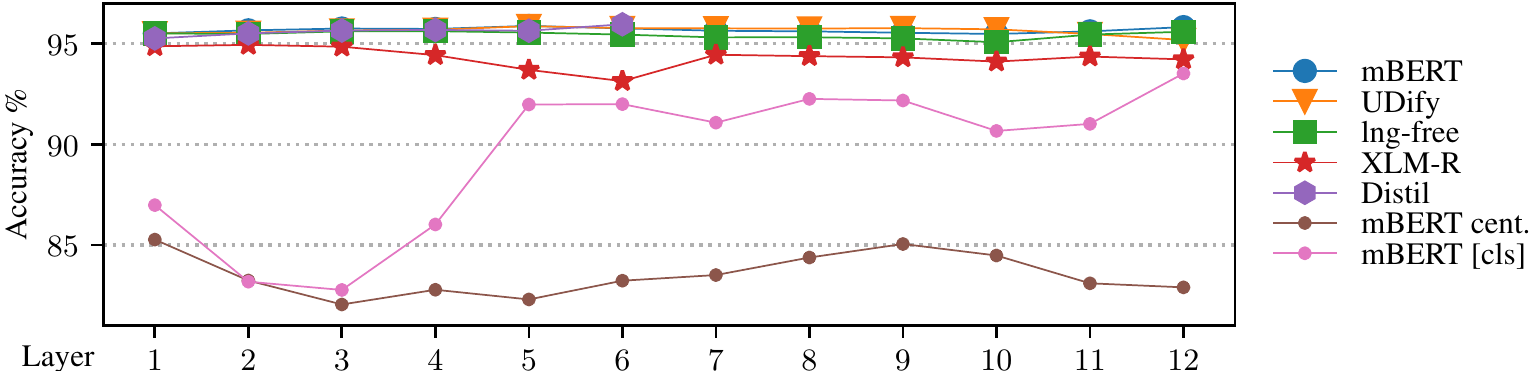}
    \caption{Language ID accuracy for different layers of
    mBERT.}\label{fig:lngid_layers}

\end{figure*}

\myparagraph{DistillmBERT.} 
This model was inferred from mBERT by knowledge distillation
\citep{sanh2019distilbert}. The model only has 6 layers instead of 12. The rest
of the hyperparameters remain the same. It was initialized with a subset of the
original mBERT parameters and trained on similar training data and optimized
towards cross-entropy of its output distribution with respect to the output of
the teacher mBERT model while keeping the MLM objective in the multi-task
learning setup.
As the model is forced to use smaller space to obtain the representation, it
might leverage the similarities between languages and reach better language
neutrality.

\myparagraph{XLM-RoBERTa.} 
\citet{conneau2019unsupervised} claim that the original mBERT is under-trained
and train a similar model on a larger dataset that consists of two terabytes of
plain text extracted from CommonCrawl \citep{wenzek2019commoncrawl}. Unlike
mBERT, XLM-R uses a SentencePiece-based vocabulary
\citep{kudo2018sentencepiece} of 250k tokens. The rest of the architecture
remains the same as in the case of mBERT\@. We train the model using the MLM
objective only, without the sentence adjacency prediction.

\begin{myfloat*}

\begin{minipage}{.68\textwidth}
\begin{figure}[H]

    \includegraphics{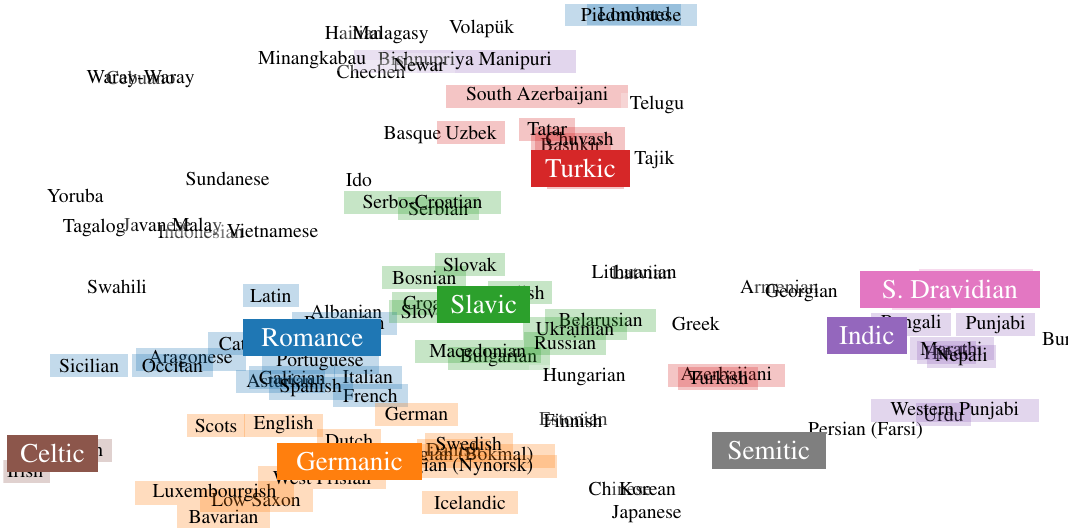}

    \caption{Language centroids of the mean-pooled representations from the 8th
    layer of cased mBERT on a tSNE plot with highlighted language
    families.}\label{fig:families}

\end{figure}
\end{minipage} \hfill%
\begin{minipage}{.30\textwidth}
\begin{table}[H]

    \centering
    \scalebox{\tablescale}{%
    \begin{tabular}{lccc}
        \toprule
                 & H    & C    & V \\ \midrule

        mBERT    & 82.0 & 82.9 & 82.4 \\
        UDify    & 80.5 & 79.7 & 80.0 \\
        lng-free & 77.1 & 80.4 & 80.6 \\
        XLM-R    & 69.7 & 69.1 & 69.3 \\
        Distil   & 81.6 & 81.1 & 81.3 \\
        random   & 60.2 & 64.3 & 62.1 \\ \bottomrule
    \end{tabular}}

    \caption{Clustering of language centroids, evaluated with homogenity,
    completened and V-Measure against genealogical language families with at
    least three mBERT languages. Averaged across layers.}\label{tab:families}

\end{table}
\end{minipage}

\end{myfloat*}

\section{Experimental Setup}

To train the language ID classifier, for each of 73 languages covered both by
mBERT and XLM-R, we randomly select 110k sentences of at least 20 characters
from Wikipedia and keep 5k for validation and 5k for testing for each language.
We also use the training data to estimate the language centroids and
training the \emph{lng-free} version of the model.

For parallel sentence retrieval, we use a multi-parallel corpus of test data
from the WMT14 evaluation campaign \citep{bojar2014findings} with 3,000
sentences in Czech, English, French, German, Hindi, and Russian.
To compute the linear projection (for the special linear projection
experimental condition), we used the WMT14 development data (500--3000
sentences per language pair).

We use manually annotated WA datasets to evaluate word alignment between
English on one side and Czech (2.5k sent.;
\citealp{marecek2016alignment})\footnote{\url{http://hdl.handle.net/11234/1-1804}},
Swedish (192 sent.;
\citealp{holmqvist2011gold})\footnote{\url{http://hdl.handle.net/11372/LRT-1517}},
German (508
sent.)\footnote{\url{https://www-i6.informatik.rwth-aachen.de/goldAlignment}},
French (447 sent.;
\citealp{och2000improved})\footnote{\url{http://web.eecs.umich.edu/~mihalcea/wpt/data/English-French.test.tar.gz}}
and Romanian (248 sent.;
\citealp{mihalcea2003evaluation})\footnote{\url{http://web.eecs.umich.edu/~mihalcea/wpt/data/Romanian-English.test.tar.gz}}
on the other side. We compare the results with FastAlign \citep{dyer2013simple}
and Efmaral \citep{ostling2016efmaral} models, which were provided with 1M
additional parallel sentences from ParaCrawl
\citep{espla2019paracrawl}\footnote{\url{https://paracrawl.eu}, Release 5}.

For MT QE, we use English-German training and test data provided for the WMT19
QE Shared Task \citep[Task~1]{fonseca2019findings}, consisting of source
sentences, automatic translations, and manually corrected reference
translations.
For the supervised estimation, we use a multilayer perceptron with a hidden
layer of size 256, trained to estimate the HTER value using the
mean-squared-error loss.

We use pre-trained tables provided by
\citet{joulin2018loss}\footnote{\url{https://fasttext.cc/docs/en/aligned-vectors.html}}
for the static word embeddings.  The embeddings were trained on Wikipedia and
aligned with a projection trained on small bilingual dictionaries. The number
of word types captured in the embedding tables spans from 350k for Romanian to
2.5M for English.

The experiments with contextualized embeddings are implemented using the
Transformers package \citep{wolf2019hugging}, which we also use for obtaining
the pre-trained models, except for UDify, which was obtained from
\citep{udify}.\footnote{\url{http://hdl.handle.net/11234/1-3042}} The
\emph{lng-free} mBERT version was finetuned using the same data that was used
for language identification.

Our source code is available at
\url{https://github.com/jlibovicky/assess-multilingual-bert}.

\section{Results}

\myparagraph{Language Identification.} 
Table~\ref{tab:lngid} and Figure~\ref{fig:lngid_layers} shows that for mBERT,
centering the sentence representations decreases the accuracy of language ID
considerably, especially in the case of mean-pooled embeddings. This result
indicates that the centering procedure indeed removes the language-specific
information to a great extent.

For comparison, the state-of-the-art language ID model from FastText
\citep{grave2018learning} reaches $91.4\%$ accuracy with a pre-trained model,
and $91.8\%$ when retrained on our training data, i.e., slightly worse than our
best model based on mBERT\@. Langid.py \citep{lui2012langid} reaches $90.1\%$
when trained on the same dataset.

Adversarial finetuning prevented the language identification only from the
\cls{} vector and only marginally for mean-pooling. This supports the
hypothesis that language identity is derived from the presence of function
words and structures and representation centering suppresses these frequent
phenomena.

%
%
Centering the representations within languages requires knowing the language in
advance. It is therefore an oracle experiment. In a sense, centering
\textit{adds} language-specific information to the representation which the
classifier might take advantage of. However, because the centering decreases
the accuracy, we can interpret this as \textit{removing} information about the
language identity.

For further comparison, we conduct the same experiment with aligned word
embeddings for 44 languages \citep{joulin2018loss}. The language ID accuracy is
99.5\% but drops to 2.3\% after centering (the same as assigning language by
chance), which supports our intuition about centering functioning as removal of
frequent patterns. Note, however, that even the experiment without centering is
an oracle experiment cannot be considered as language identification because we
need to know the language identity in advance to use the matching embeddings
table, so the accuracy is not comparable with other experiments.

\begin{table*}[t]
    \centering
    \scalebox{\tablescale}{%
    \begin{tabular}{lcccccc}
        \toprule
        & SWE & mBERT & UDify & lng-free & Distil & XLM-R \\ \midrule

        \cls{}           & ---  & .639 & .462 & .549 & .420 &  --- \\
        \cls, cent.      & ---  & .684 & .660 & .686 & .505 &  --- \\
        \cls, proj.      & ---  & .915 & .933 & .697 & .830 &  --- \\ \midrule
        mean-pool        & .113 & .776 & .314 & .755 & .600 & .883 \\
        mean-pool, cent. & .496 & .838 & .564 & .828 & .770 & .923 \\
        mean-pool, proj. & .650 & .983 & .906 & .983 & .980 & .996 \\
        \bottomrule

    \end{tabular}}
    \caption{Average accuracy for sentence retrieval over all 30 language
        pairs compared to static bilingual word embeddings (SWE).}\label{tab:retrieval}

\end{table*}
\begin{figure*}[t]

	\centering
    \includegraphics{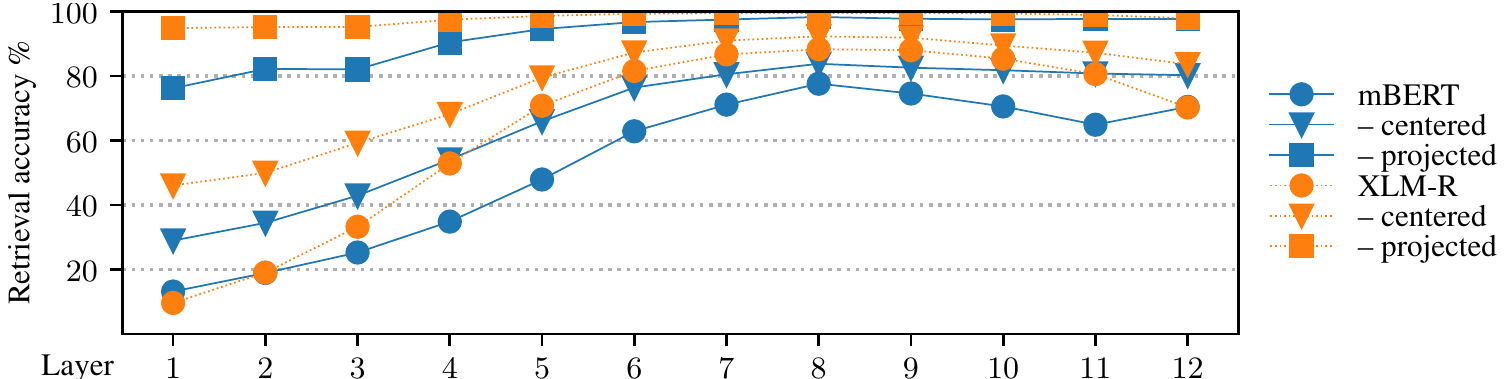}
    \caption{Accuracy of sentence retrieval for mean-pooled contextual
    embeddings from BERT layers.}\label{fig:retrieval_layers}

\end{figure*}

\myparagraph{Language Similarity.} 
Figure~\ref{fig:families} is a tSNE plot \citep{maaten2008visualizing} of the
language centroids, showing that the centroids' similarity tends to correspond
to the similarity of the languages. Table~\ref{tab:families} confirms that the
hierarchical clustering of the language centroids mostly corresponds to the
language families.

XLM-R not only preforms slightly worse in language ID, it also has worse
performance in capturing language similarity. We hypothesize that this is
because of the different approaches used in training the models.  In
particular, the next-sentence prediction used to train mBERT may lead to
stronger language-specific information because this sort of information helps
determine if two sentences are adjacent.

 \begin{table*}[h]
     \centering
     \newcommand*{\MinNumber}{0.803}%
\newcommand*{\MidNumber}{0.872} %
\newcommand*{\MaxNumber}{0.941}%

\newcommand{\ApplyGradient}[1]{%
    \IfDecimal{0#1}{%
        \ifdim #1 pt > \MidNumber pt
            \pgfmathsetmacro{\PercentColor}{max(min(100.0*(#1 - \MidNumber)/(\MaxNumber-\MidNumber),100.0),0.00)} %
            \hspace{-0.33em}\colorbox{Yellow!\PercentColor!Orange}{#1}
        \else
            \pgfmathsetmacro{\PercentColor}{max(min(100.0*(\MidNumber - #1)/(\MidNumber-\MinNumber),100.0),0.00)} %
            \hspace{-0.33em}\colorbox{Red!\PercentColor!Orange}{#1}
        \fi%
    }{#1}
}

\newcolumntype{R}{>{\collectcell\ApplyGradient}c<{\endcollectcell}}
 \setlength{\tabcolsep}{1.2pt}

\begin{tabular}{cRRRRRR}
\toprule
 & cs & de & en & es & fr & ru \\ \midrule
~cs~ & ---  & .812 & .803 & .821 & .795 & .836 \\
~de~ & .806 & ---  & .845 & .833 & .818 & .816 \\
~en~ & .783 & .834 & ---  & .863 & .860 & .809 \\
~es~ & .805 & .824 & .863 & ---  & .869 & .822 \\
~fr~ & .784 & .822 & .861 & .859 & ---  & .811 \\
~ru~ & .828 & .820 & .810 & .826 & .817 & ---  \\
\bottomrule
\end{tabular}\quad\begin{tabular}{cRRRRRR}
\toprule
 & cs & de & en & es & fr & ru \\ \midrule
~en~ & ---  & .917 & .935 & .941 & .926 & .919 \\
~cs~ & .925 & ---  & .907 & .913 & .896 & .923 \\
~de~ & .938 & .913 & ---  & .921 & .904 & .912 \\
~es~ & .936 & .907 & .916 & ---  & .934 & .908 \\
~fr~ & .928 & .903 & .917 & .935 & ---  & .905 \\
~ru~ & .920 & .910 & .918 & .910 & .903 & ---  \\
\bottomrule
\end{tabular}

     \caption{Sentence retrieval scores for the 8th layer of mBERT and XLM-R models.}\label{tab:retrievallng}

 \end{table*}

\myparagraph{Parallel Sentence Retrieval.} 
Results for mean-pooled representations in Table~\ref{tab:retrieval} reveal
that the representation centering improves the retrieval accuracy dramatically,
showing that it makes the representations more language-neutral.  An additional
50\% error reduction is achievable via learning a projection on relatively
small parallel data, leading to close-to-perfect accuracy.

Similar trends hold for all models. XLM-R significantly outperforms all models.
The UDify model that was finetuned for syntax seems to lose semantic abilities
significantly. Adversarial finetuning did not improve the performance.  The
accuracy is usually higher for mean-pooled states than for the \cls{} embedding
and varies among the languages too (see Table~\ref{tab:retrievallng}).

The accuracy also varies according to the layer of mBERT used (see
Figure~\ref{fig:retrieval_layers}). The best-performing is the 8th layer, both
for mBERT and XLM-R. These results are consistent both among models and among
tasks.

\begin{table*}[t]

    \centering
    \scalebox{\tablescale}{%
    \begin{tabular}{lcccccccc}
        \toprule
        en- & FastAlign & Efmaral & SWE & mBERT & UDify & lng-free & Distil & XLM-R \\ \midrule

        cs & .692 & .729 & .501 -- .540 & .738 & .708 & .744 & .660 & \bf .731 \\
        sv & .438 & \bf .501 & .272 -- .331 & .478 & .459 & .468 & .454 & .461 \\
        de & .741 & .759 & .473 -- .515 & .767 & .731 & \bf .768 & .723 & .762 \\
        fr & .583 & .589 & .371 -- .435 & \bf.612 & .581 & .607 & .582 & .591 \\
        ro & .690 & \bf.742 & .448 -- .470 & .703 & .696 & .704 & .669 & .732 \\

        \bottomrule

    \end{tabular}}

    \caption{Maximum F$_1$ score (usually the 8th layer) for WA across
    layers, including comparison to FastAlign and Efmaral aligners. For static word embeddings (SWE),
    we report the difference from distortion penalty introduction.}\label{tab:alignment}

\end{table*}

\begin{table*}[t]
    \centering

    \begin{tabular}{lcccccc}\toprule
    & SWE & mBERT & UDify & lng-free & Distil & XLM-R \\ \midrule

    centered             & .020 & .005 & .039 & .026 & .001 & .001 \\
    projection           & .038 & .163 & .167 & .136 & .241 & .190 \\ \midrule

    regression: SRC only & .349 & .362 & .368 & .349 & .342 & .388 \\
    regression: TGT only & .339 & .352 & .375 & .343 & .344 & .408 \\
    regression full      & .332 & .419 & .413 & .411 & .389 & .431 \\
        \bottomrule
    \end{tabular}
    \caption{Pearson correlation of estimated MT quality with HTER for
    WMT19 English-to-German translation.}\label{tab:qe}

\end{table*}

\myparagraph{Word Alignment.} 
Table~\ref{tab:alignment} shows that WA based on mBERT and XLM-R
representations match the state-of-the-art aligners trained on a large parallel
corpus. WA techniques based on multilingual contextual representations can thus
be used as a replacement of state-of-the-art statistical methods without the
use of parallel data.

The results show that the contextual embeddings well capture word-level
semantics. Furthermore, the distortion penalty does not seem to influence the
alignment quality when using the contextual embeddings, whereas for the static
word embeddings, it can make a difference of 3--6 F$_1$ points. This result
shows that the contextual embeddings encode information about the relative word
position in the sentence across languages. However, their main advantage is
still the context-awareness, which allows accurate alignment of function words.

Similarly to sentence retrieval, we experimented with explicit projection
trained on parallel data. We used an expectation-maximization approach that
alternately aligned the words and learned a linear projection between the
representations. This algorithm only brings a negligible improvement of .005
F$_1$ points.

\myparagraph{MT Quality Estimation.} 
Table~\ref{tab:qe} reveals that measuring the distance of non-centered sentence
vectors does not correlate with MT quality at all; centering or explicit
projection only leads to a mild correlation. Unlike sentence retrieval, QE is
more sensitive to subtle differences between sentences, while the projection
only seems to capture rough semantic correspondence. Note also that the Pearson
correlation used as an evaluation metric for QE might not favor the cosine
distance because semantic similarity might not linearly correspond to HTER\@.

However, supervised regression using either only the source or only MT output
shows a respectable correlation. The source sentence embedding alone can be
used for a reasonable QE\@. This means that the source sentence complexity is
already a strong indicator of the translation quality. Using the target
sentence embedding alone leads to almost as good results as using both the
source and the hypothesis, which suggests that the structure of the translation
hypothesis is what plays the important role and lexical-semantic aspects
captured by the embeddings are not sufficient for the QE\@.


The experiments with QE show that all tested contextual sentence
representations carry information about sentence difficulty for MT and
structural plausibility. However, unlike lexical-semantic features, this
information is not well accessible via simple embedding comparison.

A parallel research \citet{zhao-etal-2020-limitations,zhao2020inducing}
presents a relative success in using multilingual contextual representations
for reference-free MT evaluation. A comparison with their results suggests that
QE is a more difficult task than the reference-free MT evaluation.

\section{Conclusions}

Using a set of semantically oriented tasks, we showed that unsupervised
BERT-based multilingual contextual embeddings capture similar semantic
phenomena quite similarly across different languages. Surprisingly, in
cross-lingual semantic similarity tasks, employing cosine similarity of the
contextual embeddings without any tuning or adaptation clearly and consistently
outperforms cosine similarity of static multilingually aligned word embeddings,
even though these were explicitly trained to be language-neutral using
bilingual dictionaries.

Nevertheless, we found that vanilla contextual embeddings contain a strong
language identity signal, as demonstrated by their state-of-the-art performance
for the language identification task. We hypothesize this is due to the
sentence-adjacency objective used during training because language identity is
a strong feature for adjacency.

We explored three ways of removing the language ID from the representations in
an attempt to make them even more cross-lingual. While adversarial finetuning
of mBERT did not help, a simpler unsupervised approach of language-specific
centering of the representations managed to reach the goal to some extent,
leading to higher performance of the centered representations in the probing
tasks. The adequacy of the approach is also confirmed by a strong performance
of the computed language centroids in estimating language similarity. Still, an
even stronger language-neutrality of the representations can be achieved by
fitting a supervised linear projection on a small set of parallel sentences.

Although representation centering leads to satisfactory language neutrality, it
still requires knowing in advance what the language is. The future work thus
should focus on representations that are more language-neutral by default, not
requiring subsequent language-dependent modifications.  We hope that this work
helps to establish how future language-neutral representation should be
evaluated.

\section*{Acknowledgments}

We would like to thank Philipp Dufter and Masoud Jalili Sabet for fruitful
extensive discussions of the work.

Work done at LMU was supported by the European Research Council (ERC) under the
European Union’s Horizon 2020 research and innovation programme (grant
agreement No.~640550) and by German Research Foundation (DFG; grant FR
2829/4-1). Work done at CUNI supported by the grant 18-02196S of the Czech
Science Foundation.

\bibliography{references}
\bibliographystyle{acl_natbib}

\appendix

\section{Notes on Reproducibility}

Experiments with language identification, language similarity, and adversarial
removal of language ID were computing on GPUs. We used GeForce GTX 1080 Ti with
11GB memory. The other experiments were conducted CPUs with Intel Xeon CPU
E5--2630 v4 (2.20GHz). All experiments fitted into 32 GB RAM\@.

Models for language identification and adversarial language ID removal are
implemented in PyTorch. The linear classifier for language ID has 56k
parameters. For adversarial language ID removal, it means there are two
classifiers per layer, i.e., in total 1.3M parameters. Each experiment from
Table~\ref{tab:lngid} that includes 5 runs with different random seeds took on
average 1.38h. Results on validation data are presented in
Table~\ref{tab:lngid_valid}.

The linear projections for sentence retrieval were estimated using Scikit Learn,
which took on average 7 minutes for one model layer and one language pair,
including running the representation model in PyTorch on CPU\@. The projection
has 590k parameters. One retrieval experiment took on average 25 minutes.

We implemented the minimum weighted edge cover algorithm using the linear sum
assignment problem solver from SciPy. One experiment took on average 10
minutes.

The MT QE experiments based on cosine similarity took on average 2 minutes. The
experiments with supervised regression were trained using Scikit Learn. Each
model has 197k parameters. One experiment took on average 22 minutes.

\begin{table}[ht!]

	\centering

    \scalebox{0.52}{
    \begin{tabular}{l cc cc cc cc cc}
        \toprule
        & \multicolumn{2}{c}{mBERT}
        & \multicolumn{2}{c}{UDify}
        & \multicolumn{2}{c}{lng-free}
        & \multicolumn{2}{c}{Distil}
        & \multicolumn{2}{c}{XLM-R} \\ \midrule

        \cls{}
            & .935 & \small 12
            & .936 & \small  8
            & .798 & \small  1
            & .952 & \small  6
            & \multicolumn{2}{c}{---} \\
        \cls, cent.
            & .908 & \small 10
            & .852 & \small  8
            & .341 & \small  5
            & .825 & \small  6
            & \multicolumn{2}{c}{---} \\ \midrule
        mean-pool
            & .958 & \small  5
            B
            & .957 & \small  5
            & .956 & \small  3
            & .958 & \small  6
            & .949 & \small  1 \\
        mean-pool, cent.
            & .851 & \small  1
            & .852 & \small  1
            & .853 & \small  1
            & .841 & \small  1
            & .849 & \small  8

        \\ \bottomrule

    \end{tabular}}
    \caption{Validation accuracy of language identification for the best and
    worse scoring.}\label{tab:lngid_valid}

\end{table}

\end{document}